\documentclass[runningheads]{llncs}

\usepackage{algorithm,algorithmic}
\usepackage{amsfonts,amsmath,amssymb}
\usepackage[T1]{fontenc}
\usepackage{graphicx}
\usepackage[colorlinks=true,pdftex]{hyperref} 
\usepackage[utf8]{inputenc}
\usepackage{listings}
\usepackage{url}
\usepackage{wrapfig} 
\usepackage{caption}
\usepackage{subcaption}
\usepackage{placeins}
\usepackage{comment}

\usepackage{pifont}
\usepackage{multirow}
\usepackage{array}

\usepackage{color}
\usepackage{xcolor}

\newcommand{\KILL}[1]{}

\begin{document}
\definecolor{darkgreen}{RGB}{0,128,0}
\newcommand{\cmark}{\textcolor{darkgreen}{\ding{51}}} 
\newcommand{\xmark}{\textcolor{red}{\ding{55}}} 

\title{OntView: What you See is What you Meant}
%

\author{Carlos Bobed\inst{1,2} \and 
 Carlota Quintana\inst{1} \and 
 Eduardo Mena\inst{1,2} \and 
 Jorge Bobed\inst{3} \and
 Fernando Bobillo\inst{1,2} 
}

\authorrunning{C. Bobed et al.}

\institute{
University of Zaragoza, Zaragoza, Spain
\and
Aragon Institute of Engineering Research (I3A), Zaragoza, Spain 
\and 
NTT Data\\
\email{\{cbobed,cquintana,emena,fbobillo\}@unizar.es, jorge.bobed.lisbona@nttdata.com}
}

\maketitle


\begin{abstract}

In the field of knowledge management and computer science, ontologies provide a structured framework for modeling domain-specific knowledge by defining concepts and their relationships. However, the lack of tools that provide effective visualization is still a significant challenge. While numerous ontology editors and viewers exist, most of them fail to graphically represent ontology structures in a meaningful and non-overwhelming way, limiting users' ability to comprehend dependencies and properties within 
large ontological frameworks. 

In this paper, we present OntView, an ontology viewer that is designed to provide users with an intuitive visual representation of ontology concepts and their formal definitions through a user-friendly interface. Building on the use of a DL reasoner, OntView follows a ``What you see is what you meant'' paradigm, showing the actual inferred knowledge. One key aspect for this is its ability to visualize General Concept Inclusions (GCI), a feature absent in existing visualization tools. Moreover, to avoid a possible information overload, Ontview also offers different ways to show a simplified view of the ontology by:~1)~creating ontology summaries by assessing the importance of the concepts (according to different available algorithms),~2)~focusing the visualization on the existing TBox elements between two given classes 
and~3)~allowing to hide/show different branches in a dynamic way without losing the semantics. 
OntView has been released with an open-source license for the whole community.   

\keywords{Ontologies  \and Visualization \and Description Logics}

\end{abstract}


\section{Introduction} 
\label{sec:intro}


Ontologies are fundamental to the Semantic Web as they provide a framework for organizing knowledge by defining concepts and relationships within a specific domain in a  
structured and machine-readable way, 
helping computer systems to generate new insights, identify inconsistencies, and classify data automatically.

One of the challenges in the Semantic Web field is the effective visualization of ontologies, to aid users in understanding and interacting with ontological structures. Being able to visualize ontologies effectively is essential for users to fully take advantage of the potential that these frameworks offer: an appropriate graphical representation of ontologies is essential for users to explore, analyze, and understand the ontology elements. 
For example, ontology visualization helps both knowledge engineers to identify potential modeling errors (inconsistencies), and developers to understand and reuse ontologies created by others. 
However, as ontologies grow larger and more complex, visualizing them in a clear and manageable way becomes increasingly difficult due, among others, to issues such as information overload. Moreover, for the proper modeling and use of an ontology, a deep understanding of the concepts and relationships to be represented is required, as well as knowledge of the complexity of Description Logic (DL) formalisms~\cite{DLHandbook}. 

Although several ontology visualization tools have been developed, the effective visualization of ontologies is still an open problem. In this paper, we present OntView, a novel ontology viewer which specifically addresses these needs by providing an intuitive and interactive interface that not only aids in visualizing ontologies but also supports error detection, enhances understanding, and allows seamless navigation through complex and large knowledge structures. Ontview is publicly available\footnote{\url{https://sid.cps.unizar.es/projects/OntView/}, last accessed June 20, 2025.} and its key features can be summarized as follows: 

\begin{itemize}
\item \emph{Meaningful Semantic Visualization}: OntView represents ontologies following the ``what you see is what you meant'' paradigm, using a rich and expressive visual language, with the help of a Description Logics reasoner. Anonymous classes and  descriptions, as well as General Concept Inclusions (GCIs) are not left behind, and are presented to provide the required view to apprehend the actual semantics. This allows a graphical representation of all the ontology abstractions/expressions whether they have a name or not.

\item \emph{Interactive Navigation}: OntView allows users to explore the ontology hiding and expanding the different elements in an interactive way. OntView keeps track of the actual reasoned model 
so that displaying elements that were hidden is fast.
Besides, the user is continually provided with information about the size of the hidden/visible fragments of the ontology to have a broader view on the whole domain. Moreover, OntView allows users to control the incremental exploration of the ontology graph by specifying the amount of nodes that the user wants to expand at each step. Finally, users can search for ontology terms anytime.

\item \emph{Information Overload Counter-measures}: As displaying large ontologies (with hundreds or thousands of terms) turns soon overwhelming for the user, OntView is able to generate a simplified ontology view automatically (via concept relevance measures) at a customized level of detail, which summarizes the ontology without losing semantics, i.e., displaying the most important elements. Moreover, OntView allows users to define fragments of the ontology for which anonymous classes are displayed.

\end{itemize}

The remainder of this paper is organized as follows. Firstly, Section~\ref{sec:relatedWork} compares OntView with the most relevant existing ontology viewers. Then, Section~\ref{sec:visualLang} details the expressive visual language used by OntView to present the different ontology elements. Section~\ref{sec:processing} details the techniques used to visualize the GCIs and anonymous classes, and the different
methods to summarize the view of an ontology. Section~\ref{sec:evaluation} discusses some implementation notes, and, finally, Section~\ref{sec:conclusions} sets out some conclusions and ideas for future research.


\section{Related Work}
\label{sec:relatedWork}

While there have been a plethora of visualization proposals, this particular issue is still open (as the Voila! workshop series\footnote{\url{https://voila-workshop.github.io/}, last accessed May 14, 2025.} illustrates, co-located with ISWC along nine editions). In this section, we will focus on the most currently relevant ontology visualization projects, referring the interested reader to the two main surveys on the subject carried out by Katifory~et~al.~\cite{Katifori2007} and Dudavs~et~al.~\cite{Dudavs2018}. We leave aside the works on Linked Data visualization for the time being as our main goal is to ease the schema understanding process, referring this time the interested reader to Dadzie~et~al.~\cite{Dadzie2011} and Po~et~al.~\cite{Po2020} for further related works. 

Thus, we have selected OWLViz\footnote{\url{https://github.com/protegeproject/owlviz}, last accessed May 14, 2025.}, OntoGraf\footnote{\url{https://github.com/protegeproject/ontograf}, last accessed May 14, 2025.}, WebVOWL~\cite{Janowicz2016}, OWLGrED~\cite{Barzdicnvs2010}, and KC-Viz~\cite{motta2011} as the most notable related (and open) works. It is important to note that OntoGraf and OWLViz are plugins for Protégé~\cite{Musen2015}, a well-known and widely used ontology editing tool. These plugins rely on Protégé for their functionality and cannot operate independently without it, as they are tightly integrated into Protégé’s environment.

\emph{OWLViz} is a plugin that clearly displays hierarchical levels and distinguishes between asserted and inferred data, with options to show or hide nodes. However, it cannot visualize properties, instances, or anonymous classes. Furthermore, it only offers a static (non‐interactive) view and lacks features to summarize or display the entire ontology at once, making it hard to navigate large ontologies.

\emph{OntoGraf} is an advanced plugin for visualizing ontologies, featuring customizable layouts (cube, tree, or vertical tree), node hiding/display, export options, and differentiation between asserted and inferred knowledge. However, it has a steep learning curve due to its complex connectors, non-intuitive node expansion (one-by-one or via wildcard), and reliance on Protégé expertise.

\emph{WebVOWL} is a tool based on the VOWL visual language~\cite{Janowicz2016} for representing OWL ontologies. Its strengths include a detailed information panel that appears when selecting a node, displaying attributes such as name, type, domain, and range. It also offers search functionality for large ontologies, experimental editing features, and export options to formats such as URL, JSON, SVG, TeX, and TTL. However, it has shortcomings in visualizing hierarchical levels, arranges nodes confusingly (sometimes duplicating them), causes property labels to overlap, and, when the ontology grows, nodes overlap as well, reducing overall legibility.

\emph{OWLGrEd} is an OWL ontology visualization and editing tool that uses UML-style diagrams, allowing direct in-diagram editing, export to OWL, SVG, and image formats, and customizable node styling. However, its diagrams can be confusing, it lacks reasoning support and partial-view capability, and it does not display hierarchical levels.

\emph{KC-Viz} is a NeOn Toolkit plugin that extracts and displays key concepts to deliver concise, middle-out summaries of large ontologies, offering incremental subtree expansion/hiding, adjustable summary size, layout control, history browsing, seamless integration with the Ontology Navigator and Entity Properties View, and save/load of custom views (both taxonomic and domain–range). However, its tree layout cannot rotate (often causing label overlap), and while they used one of the key-concept extraction algorithms that we have adopted (KCE~\cite{Peroni2008}), it was not possible to set any user-tuning options. Although KC-Viz is not currently available, we have made an exception to include it in our analysis because it is the conceptually closest ontology viewer to OntView. However, OntView extends the semantic features of the visualization provided.

\begin{table}[!htb]
\begin{scriptsize}
\begin{center}
\begin{tabular}{|r|c|c|c|c|c|c|}
\hline
 & OWLViz & OntoGraf & WebVOWL & OWLGrED & KC-Viz & OntView \\ \hline \hline
Display asserted graph & \cmark & \cmark & \cmark & \cmark & \cmark & \xmark \\ \hline
Display inferred graph & \cmark & \cmark & \xmark & \xmark & \xmark &  \cmark \\  \hline
Reasoner support  &  \cmark & \cmark  & \xmark & \xmark & \xmark & \cmark \\ \hline \hline
Visual class hierarchy  & \cmark  & \cmark  & \xmark & \xmark & \cmark & \cmark \\ \hline
View role/property hierarchies  & \xmark  & \xmark & \cmark  & \cmark & \xmark & \cmark \\ \hline
Display anonymous classes & \xmark & \xmark & \xmark & \xmark & \xmark &  \cmark \\ \hline
Display GCIs  & \xmark  & \xmark & \xmark  & \xmark & \xmark & \cmark \\ \hline
Control detail level  & \xmark & \xmark  & \xmark & \xmark & \xmark  & \cmark \\ \hline
List class instances & \xmark  & \xmark & \cmark & \cmark & \xmark & \cmark \\ \hline \hline
Modify elements layout on screen & \xmark & \cmark & \cmark & \cmark & \cmark & \cmark \\ \hline
Incremental navigation  & \xmark & \xmark  & \cmark & \xmark & \cmark & \cmark \\ \hline
Partial graph hiding  & \cmark & \cmark & \cmark & \xmark & \cmark &  \cmark \\ \hline \hline
Zoom  & \cmark  & \cmark & \cmark  & \cmark & \cmark & \cmark \\ \hline
Diagram overview navigation  & \xmark  & \xmark & \xmark  & \cmark & \xmark & \xmark \\ \hline
Save current state  & \xmark  & \cmark & \cmark  & \cmark & \cmark & \cmark \\ \hline
Export images  & \cmark & \cmark  & \xmark & \xmark & \xmark & \cmark \\ \hline
\end{tabular}
\end{center}
\end{scriptsize}
\caption{Comparison of ontology viewers}
\label{tabla:comparacion2}
\end{table}

Table~\ref{tabla:comparacion2} summarizes the features of each ontology viewer and compares them with OntView. 
In summary, no single visualization tool excels across all criteria: OWLViz and OntoGraf provide strong support for exploring asserted and inferred hierarchies within the Protégé environment, whereas WebVOWL and OWLGrEd offer rich role/property views and export capabilities but falter on hierarchy clarity. KC-Viz's innovative key-concept summarization remains a valuable reference despite its current non-functionality. This comparison underscores the ongoing challenge of balancing comprehensive ontology views with interactive, scalable layouts; a challenge that OntView aims to address through an independent, feature-rich visualization framework. Furthermore, OntView provides some unique features: showing anonymous classes and GCIs, and allowing users to control the  level of detail. 

Finally, due to space limitations, we provide some additional graphical material\footnote{\url{http://sid.cps.unizar.es/projects/OntView/Additional/}, last accessed May 14, 2025.} to ease the comparison among the different viewers included in this related work when they present the same ontology.


\section{Visual Language}
\label{sec:visualLang}


In OntView, we have adopted the ``what you see is what you meant'' motto, using a Description Logic reasoner to obtain the actual classified hierarchy and place every ontological element and definition accordingly. We adopt a level order to capture the subsumption relationships visually: more general classes are always on the left-most levels, and the user gets more specialized classes as s/he moves to the right-most levels. 

\begin{figure}[!htb]
    \centering
    \includegraphics[width=1\textwidth]{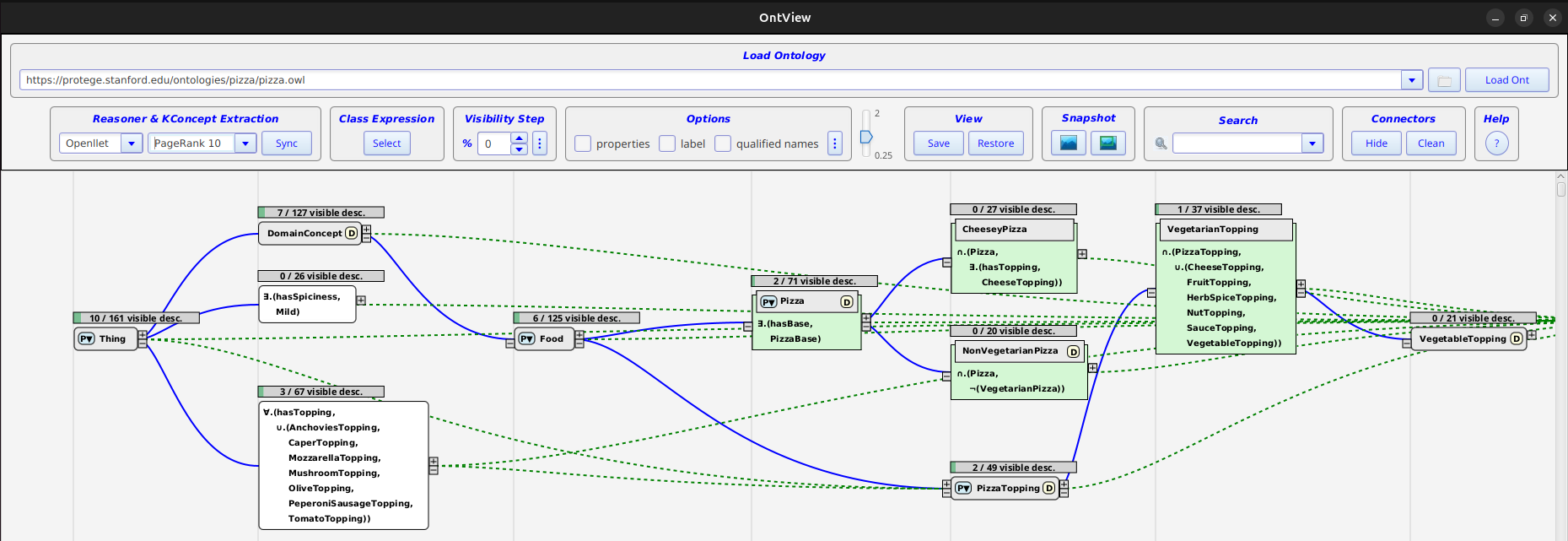}
    \caption{OntView GUI.}
    \label{fig:GUI_conOnt}
\end{figure}

Figure~\ref{fig:GUI_conOnt} shows the viewer's graphical user interface (GUI) with a loaded ontology, highlighting the functionalities accessible at the top of the screen. To illustrate OntView, we chose to use Pizza ontology\footnote{\url{https://protege.stanford.edu/ontologies/pizza/pizza.owl}, lat accessed May 14, 2025.}, which is well-known by many potential readers. 

\begin{figure}[!htb]
    \centering
    \begin{subfigure}[c]{0.61\textwidth}
        \centering
        \includegraphics[width=\textwidth]{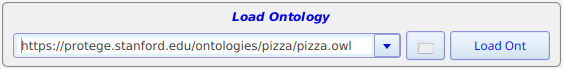}
        \caption{}
        \label{fig:load}
    \end{subfigure}%
    ~ 
    \begin{subfigure}[c]{0.33\textwidth}
        \centering
        \includegraphics[width=\textwidth]{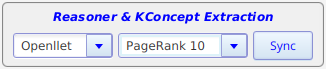}
        \caption{}
        \label{fig:KCEwithOption}
    \end{subfigure}
    ~ 
    \begin{subfigure}[c]{0.18\textwidth}
        \centering
        \includegraphics[width=\textwidth]{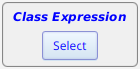}
        \caption{}
        \label{fig:cExpression}
    \end{subfigure}
    ~
    \begin{subfigure}[c]{0.165\textwidth}
        \centering
        \includegraphics[width=\textwidth]{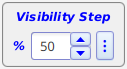}
        \caption{}
        \label{fig:visibilityStep}
    \end{subfigure}
    ~
    \begin{subfigure}[c]{0.41\textwidth}
        \centering
        \includegraphics[width=\textwidth]{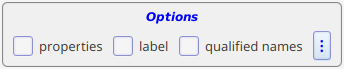}
        \caption{}
        \label{fig:options}
    \end{subfigure}
    ~
    \begin{subfigure}[c]{0.075\textwidth}
        \centering
        \includegraphics[width=\textwidth]{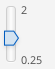}
        \caption{}
        \label{fig:zoom}
    \end{subfigure}
    ~
    \begin{subfigure}[c]{0.185\textwidth}
        \centering
        \includegraphics[width=\textwidth]{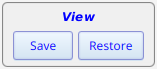}
        \caption{}
        \label{fig:view}
    \end{subfigure}
    ~
    \begin{subfigure}[c]{0.135\textwidth}
        \centering
        \includegraphics[width=\textwidth]{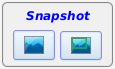}
        \caption{}
        \label{fig:snapchot}
    \end{subfigure}
    ~
    \begin{subfigure}[c]{0.29\textwidth}
        \centering
        \includegraphics[width=\textwidth]{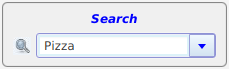}
        \caption{}
        \label{fig:search}
    \end{subfigure}
    ~
    \begin{subfigure}[c]{0.19\textwidth}
        \centering
        \includegraphics[width=\textwidth]{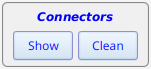}
        \caption{}
        \label{fig:connectors}
    \end{subfigure}
    ~
    \begin{subfigure}[c]{0.06\textwidth}
        \centering
        \includegraphics[width=\textwidth]{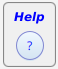}
        \caption{}
        \label{fig:help}
    \end{subfigure}
    \caption{OntView menu: a) Load ontology, b) Reasoner \& Key Concept Extractor, c) Class expression, d) Visibility step, e) Visualization options, f) Zoom, g) View, h) Snapshot, i) Search node, j) Show/Clean connectors, k) Help}
    \label{fig:tools}
\end{figure}

Figure~\ref{fig:tools} shows the available options with a higher level of detail. They include ontology loading (``Load Ontology''), ontology reasoner and concept extractor technique selection (``Reasoner \& KConcept Extraction'', see Section~\ref{subsec:summaries}), 
class expression selection (``class expression'', see Section~\ref{subsec:harvesting}), expand/collapse visibility by percentage step (``Visibility step'', see Section~\ref{subsec:summaries}), visualization options selection (deciding whether to show properties, labels, and qualified names) (``Options''), zoom level, saving/restoring views (``View''), saving screenshots of the current view/ontology as PNG image (``Snapshot''), search of primitive, defined, or anonymous classes (``Search''), connector showing/hiding (``Connectors''), and help (``Help'').

\begin{figure}[!htb]
    \centering
    \begin{subfigure}[c]{0.3\textwidth}
        \centering
        \includegraphics[width=\textwidth]{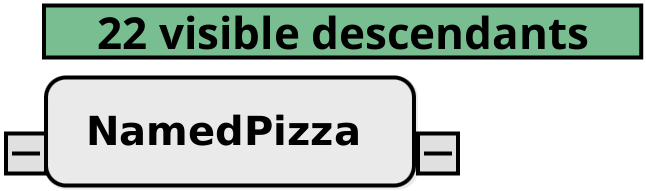}
        \caption{}
        \label{fig:namedClasses}
    \end{subfigure}%
    ~ 
    \begin{subfigure}[c]{0.3\textwidth}
        \centering
        \includegraphics[width=\textwidth]{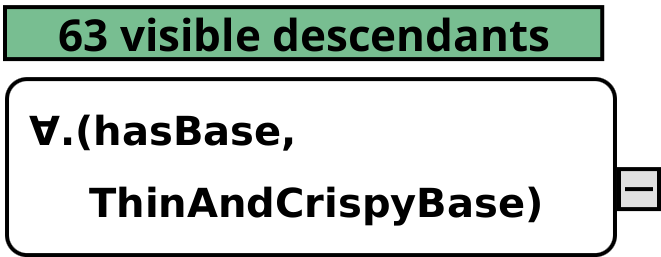}
        \caption{}
        \label{fig:anonymousClasses}
    \end{subfigure}
    ~
    \begin{subfigure}[c]{0.25\textwidth}
        \centering
        \includegraphics[width=\textwidth]{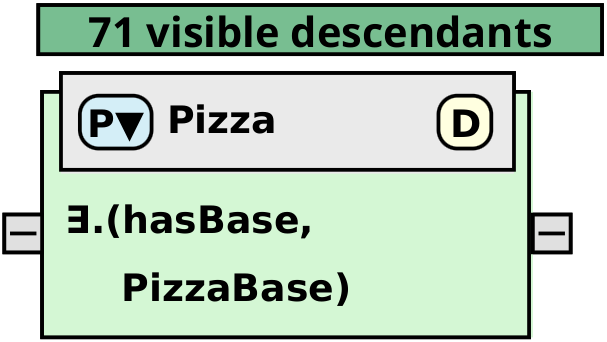}
        \caption{}
        \label{fig:equivalentClasses}
    \end{subfigure}
    \caption{Classes representation: a) Primitive classes, b) Anonymous classes, and c) Defined and equivalent classes.}
    \label{fig:classRep}
\end{figure}

The graphical elements of the OntView viewer represent classes and relationships, organized in levels representing their hierarchies within the ontology. The main key visual elements include the following ones:

\noindent\textit{Primitive Classes:} Basic concepts forming the ontology's vocabulary are displayed as light gray rectangles with visible names (Fig.~\ref{fig:namedClasses}).

\noindent\textit{Anonymous Classes:} They include concepts without explicit names in the schema, used in the ontology as part of complex concept expressions, e.g., in a concept equivalence axiom, often representing sets of elements (Fig.~\ref{fig:anonymousClasses}).

\noindent\textit{Defined or Equivalent Classes:} They represent concepts specified with necessary and sufficient conditions for membership. OntView selects an atomic concept (\texttt{OWLClass}) from the equivalent ones to be displayed as main label, 
and highlights the rest of equivalent classes and class expressions in a green rectangle below~(Fig.~\ref{fig:equivalentClasses}).

\noindent For all the types of class nodes in the graph, OntView keeps track of the total amount of children nodes and the amount of visible ones to provide the user with a broad view of the underlying size of the ontology under such nodes. This visual feedback is represented in the upper bar that each node has, showing the ratio of currently visible children. The [-] handles shown in the sides of class nodes allow to show/hide their parent and children nodes; the green bar on top of each class node informs the user also about the absolute number of its visible descendants, which can be taken into account when interacting with that node.

\noindent
\begin{minipage}{0.7\textwidth}
    \noindent\textit{Class Disjointness:} This is indicated by a small ``\texttt{D}'' marker on the corresponding concepts. Clicking the marker reveals the disjoint relationships. These are also depicted as two red lines, indicating that the connected concepts cannot have shared instances (Fig.~\ref{fig:disjoints}).
\end{minipage}%
\hspace{0.03\textwidth}
\begin{minipage}{0.3\textwidth}
    \centering
    \includegraphics[width=0.7\textwidth]{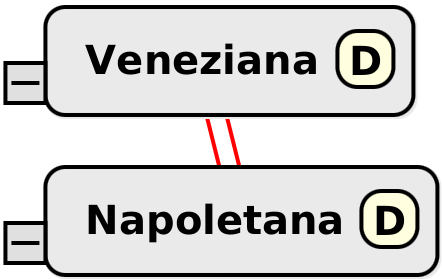} 
    \captionof{figure}{Disjoint classes.}
    \label{fig:disjoints}
\end{minipage}

\noindent
\begin{minipage}{0.7\textwidth}
    \noindent\textit{Object and Data Properties:} OntView marks with a small ``\texttt{P}''  symbol inside the graph node those class expressions that are asserted as the domain of one or more properties (Fig~\ref{fig:propiedadesP}). Clicking on the marker deploys a list of properties (both Object and Data Properties) beneath the graph node. 
\end{minipage}%
\hspace{0.03\textwidth}
\begin{minipage}{0.3\textwidth}
    \centering
    \includegraphics[width=0.9\textwidth]{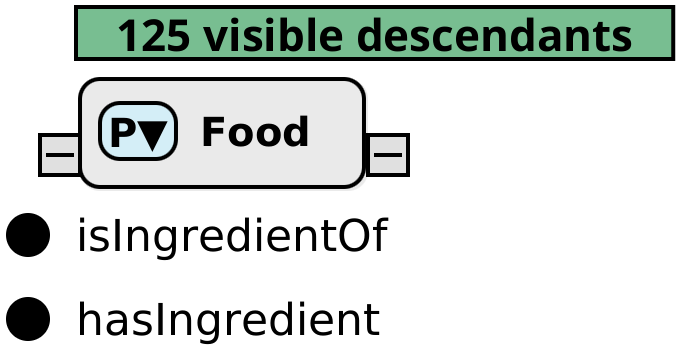} 
    \captionof{figure}{Properties.}
    \label{fig:propiedadesP}
\end{minipage}


~

\begin{figure}[!htb]
    \centering
    \includegraphics[width=\textwidth]{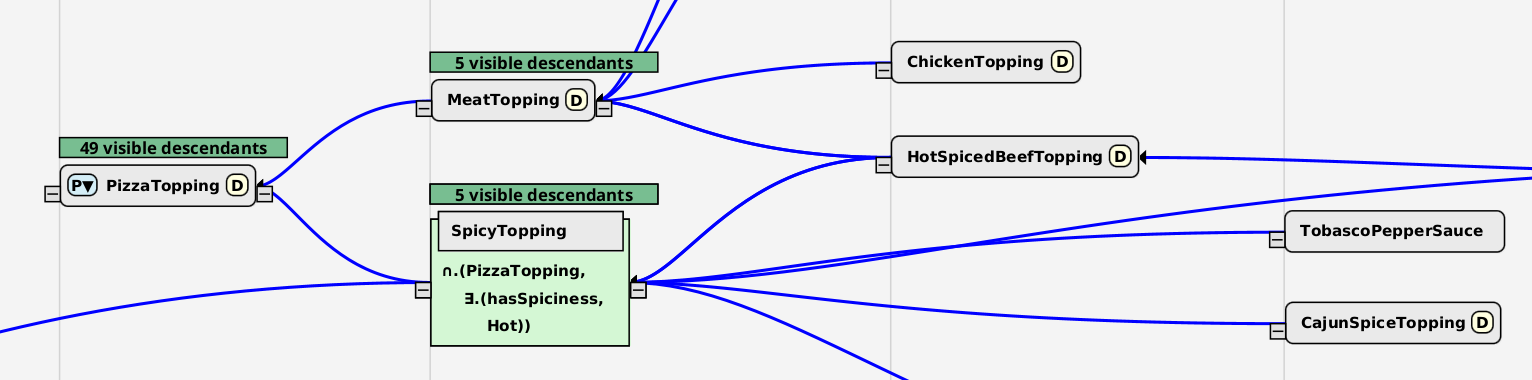}
    \caption{Graph levels in OntView: Each node always have all its parents at left-most levels and its descendants at right-most ones.}
    \label{fig:level}
\end{figure}

\noindent\textit{Graph Levels:}
The graph is organized into horizontal levels corresponding to the maximum hierarchical distance of a concept from \texttt{OwlThing} (Fig.~\ref{fig:level}). The light gray vertical lines in the background visually mark these levels, helping to distinguish and structure the hierarchical relationships between classes even if the subsumption connectors are not displayed. To establish which level each class belongs to, we recursively calculate the level of all its direct superclasses and we assign the next one (i.e., a class is always displayed at a deeper level than all its ancestors). Establishing this level-based visual constraint aims at helping the users in their interpretation of the knowledge.

\begin{figure}[H]
    \centering
    \begin{subfigure}[c]{0.85\textwidth}
        \centering
        \includegraphics[width=\textwidth]{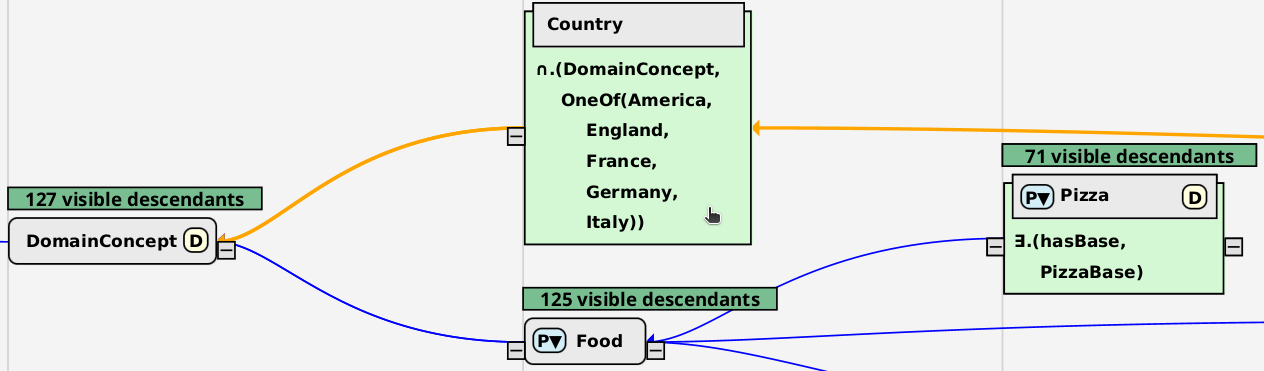}
        \caption{}
        \label{fig:isAConncectors}
    \end{subfigure}
    ~
    \begin{subfigure}[c]{0.7\textwidth}
        \centering
        \includegraphics[width=\textwidth]{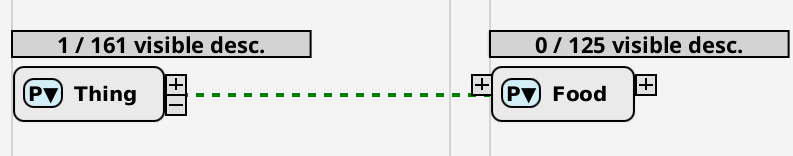}
        \caption{}
        \label{fig:dashedConnectors}
    \end{subfigure}
    \caption{Subsumption relationships: a) IsA Connectors, b) Dashed Connectors presenting hidden subsumption relationships.}
    \label{fig:classRep}
\end{figure}

\noindent\textit{IsA Connectors:} 
These connectors represent direct hierarchical relationships between concepts in the graph. They are drawn as solid lines, typically blue, which turn orange when a specific node is selected. An IsA connector indicates that a concept is a subclass of another one, establishing a ``is-a'' relationship from right to left (Fig.~\ref{fig:isAConncectors}).

\noindent\textit{Dashed Connectors:} A special IsA connector, shown as a dashed line, represents an indirect hierarchical relationship (i.e., there exist intermediate nodes that are currently not visible/displayed) preserving the visual relationship between the visible nodes (Fig.~\ref{fig:dashedConnectors}). 

\begin{figure}[htbp]
    \centering
    \includegraphics[width=0.9\textwidth]{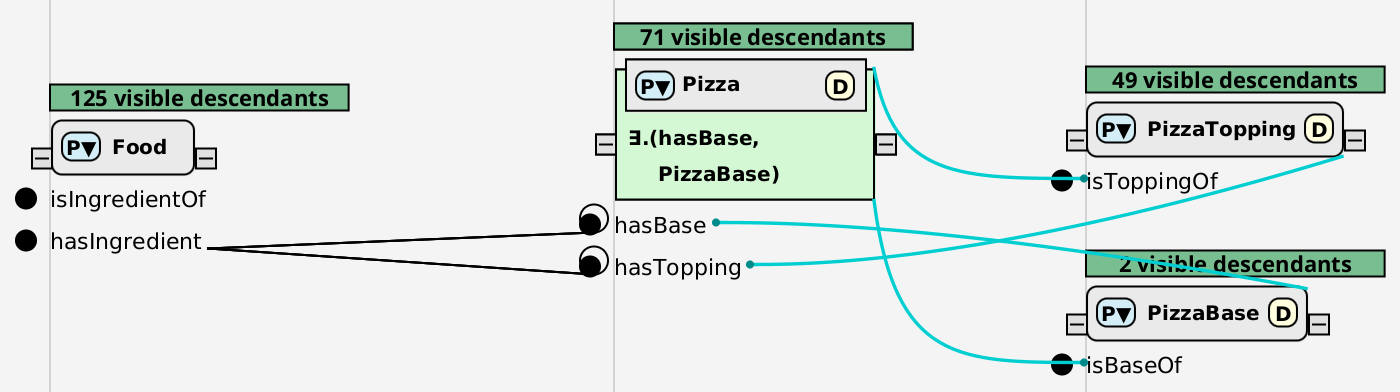}
    \caption{Range (light blue) and SubProperty (pink) connectors.}
    \label{fig:rango}
\end{figure}

\noindent\textit{Range Connectors:}
While properties are placed beneath their domain nodes, the ranges of object properties are defined by a link to a particular node (see light blue connectors in Fig.~\ref{fig:rango}). For example, \textit{hasBase} has the range \textit{PizzaBase}. In data properties, the range datatype is shown (e.g., \texttt{xsd:string}). Ontview also tries to minimize line crossings as much as possible.

\noindent\textit{Property Hierarchy Connectors:} Object Property hierarchies are also rendered in OntView via inheritance connectors showing their subProperty relationship. They are represented by a black line connecting the subproperty to the parent property from right to left (see black connectors in Fig.~\ref{fig:rango}). 

Due to space limitations, we have focused here on the most relevant visual elements; however, all these elements are complemented with different tooltips and contextual menus to show all the information inferred about each ontological element, such as the ABox extension of the different classes or further information about the properties (e.g., functional, inverse properties, transitivity, \ldots). 


\section{Advanced Ontology Visualization Techniques}
\label{sec:processing}

Two of the distinguishing features of OntView are the capability of visualizing the GCIs and anonymous classes, as well as the possibility of adding different methods to summarize the view on the ontology. In this section, we detail how we have implemented both features in an efficient way. 

\subsection{Processing GCIs and Anonymous Classes}
\label{subsec:harvesting}

Adopting expressive DLs, such as the logic on which the OWL~2 language is based, allows to model complex knowledge, which comes captured by complex class expressions and GCI axioms. Such expressions and axioms are not usually displayed in existing ontology viewers. 

Unfortunately, when working with the class hierarchy using directly a DL reasoner through the OWLAPI, these important elements are hidden\footnote{For example, \texttt{getSubClasses} or \texttt{get*PropertyDomains} methods of \texttt{OWLReasoner} return a \texttt{NodeSet<OWLClass>}, only named classes.}. Thus, in order to make them accessible, our first approach was to rewrite the ontology via (semantic) emulation~\cite{Krotzsch2015}: we performed a \emph{structural reduction} by naming all the anonymous classes introducing defined fresh concepts, thus forming a conservative extension of the original ontology (i.e., not changing their satisfying models). Such fresh symbols acted as representatives for the expressions that otherwise were not actionable via the \texttt{OWLReasoner} class of the OWLAPI. However, our empirically evaluations showed that this came at the cost of putting too high extra computation requirements to the DL reasoner.

Therefore, we opted for a different approach. Rather than classifying all the anonymous classes, which can be unfeasible for large expressive ontologies, we only classify those anonymous classes that will be presented in the user GUI. Apart from gathering all the anonymous class expressions in the ontology as before, we performed a two-step graph building algorithm: 

\begin{enumerate}

    \item We build an initial hierarchy graph only with the named classes and their isA relationships inferred from the ontology (thus, accessible both from \texttt{OWLOntology}'s and \texttt{OWLReasoner}'s methods). This initial graph acts as a scaffold to place the gathered anonymous expressions where required.
    
    \item We find the location of each anonymous class expression appearing in the ontology axioms 
    by performing subsumption tests with the DL reasoner. 
    As all the expressions are already considered when classifying the ontology, we can assume that there are not inconsistencies in the ontology, and then, we calculate the direct superclasses of each expression class. Note that we incrementally do this, pushing monotonically each expression to its location in the hierarchy. At the same time, we check for equivalences which might not be directly defined. 
\end{enumerate}

This approach allows us to create all the required nodes for the anonymous class expressions in the ontology. For example, if the domain of a property $R$ is a complex class expression $A \sqcap (\exists S.\top)$, OntView creates a node representing such an expression and the object property $R$ is linked in the node.

\paragraph{Controlling the Detail Level.} 

\begin{figure}[H]
    \centering
    \includegraphics[width=0.9\textwidth]{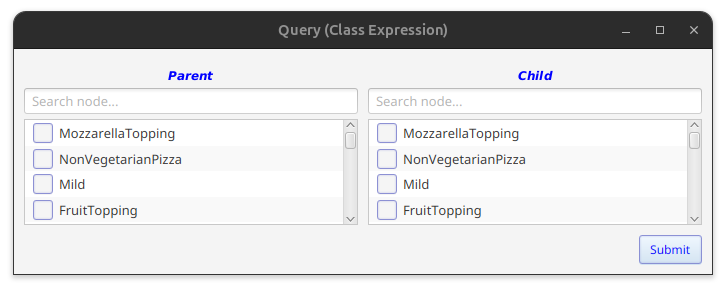}
    \caption{GUI to select the classes to control the level of detail.}
    \label{fig:classExpression}
\end{figure}

As the amount of anonymous expressions and GCIs can be quite huge in large ontologies, we have implemented a control mechanism to let the user decide which part of the ontology should be further detailed. We allow the user to define the fragment of the concept taxonomy to be detailed by selecting two concepts $\mathcal{C}_1$ and $\mathcal{C}_2$ such that $\mathcal{C}_2 \sqsubseteq \mathcal{C}_1$. By default, $\mathcal{C}_1$ is $\top$ and $\mathcal{C}_2$ is $\bot$. 
Then, OntView only inserts in the graph  those anonymous class expressions $\mathcal{CE}_i$ such that $\mathcal{C}_2 \sqsubseteq \mathcal{CE}_i \sqsubseteq \mathcal{C}_1$ holds. We keep the ontology materialized in memory (via an \texttt{OWLReasoner} instance with all the possible precomputed inferences) to speed this step up. The interface of this feature can be seen in Fig.~\ref{fig:classExpression}.

\subsection{Summarizing the Ontology for Visual Purposes}
\label{subsec:summaries}

 Visual space and user's attention are scarce resources. Indeed, showing the whole ontology at once is nice, but it is not always useful. Thus, inspired by another adaptive knowledge approach~\cite{ESWA2024}, we have adopted an approach based on selecting the most relevant concepts in the ontology to build an ontology view on top of them. OntView exploits different relevance algorithms to assess the importance of the different concepts and select which nodes are to be shown initially. This mechanism is implemented in a modular way, and can be easily extended to give your own approach. 

In the current implementation, we provide three main summarizing methods based on:


\begin{itemize}
\item KCE $n$ (Key Concept Extraction techniques as proposed by Peroni et al.~\cite{Peroni2008}): This method measures the importance of the different concepts according to a series of cognitive, statistical and topological measures\footnote{The original source code can be found in \url{https://github.com/essepuntato/KCE}, last accessed May 14, 2025.} and retrieves the $n$ most relevant ones\footnote{The amount of key concepts in this method acts as a lower bound as the API, when they have the same score, returns all the tied concepts.}. The only drawback is that, for the time being, it does only consider named concepts and an adaptation of some of the inner metrics should be considered to include anonymous classes in the mix\footnote{In previous versions of OntView (with the structural reduction presented in the previous section), we used the fresh symbols also as handles for this relevance assessment. However, they affected some of the inner metrics and, besides, introducing more named concepts incurred in an extra computational overhead.}.

\item PageRank/RDFRank $n$: graph centrality measures applied to the class taxonomy. Both exploit the PageRank algorithm over RDF triples~\cite{diefenbach2018}, with their difference being whether they consider the graph as a directed (pure PageRank) or an undirected (or bidirectional) one. These methods allow to consider as well the relevance of anonymous classes.

\item ``Do your own summary'': sometimes the user is just interested on seeing how a set of particular concepts might be related within the ontology. OntView makes this possible by allowing to select the concepts to be displayed through a dialog which can be seen in Fig.~\ref{fig:CustomKCE}.
\begin{figure}[H]
    \centering
    \includegraphics[width=0.9\textwidth]{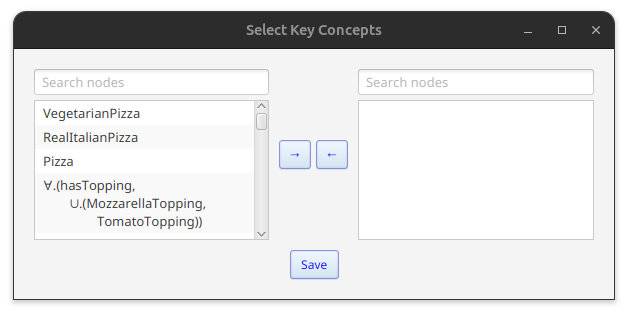}
    \caption{GUI to select the concepts for our Custom Key Concept Extractor.}
    \label{fig:CustomKCE}
\end{figure}
\end{itemize}

\begin{figure}[!htb]
    \centering
    \begin{subfigure}[c]{0.7\textwidth}
        \centering
        \includegraphics[width=\textwidth]{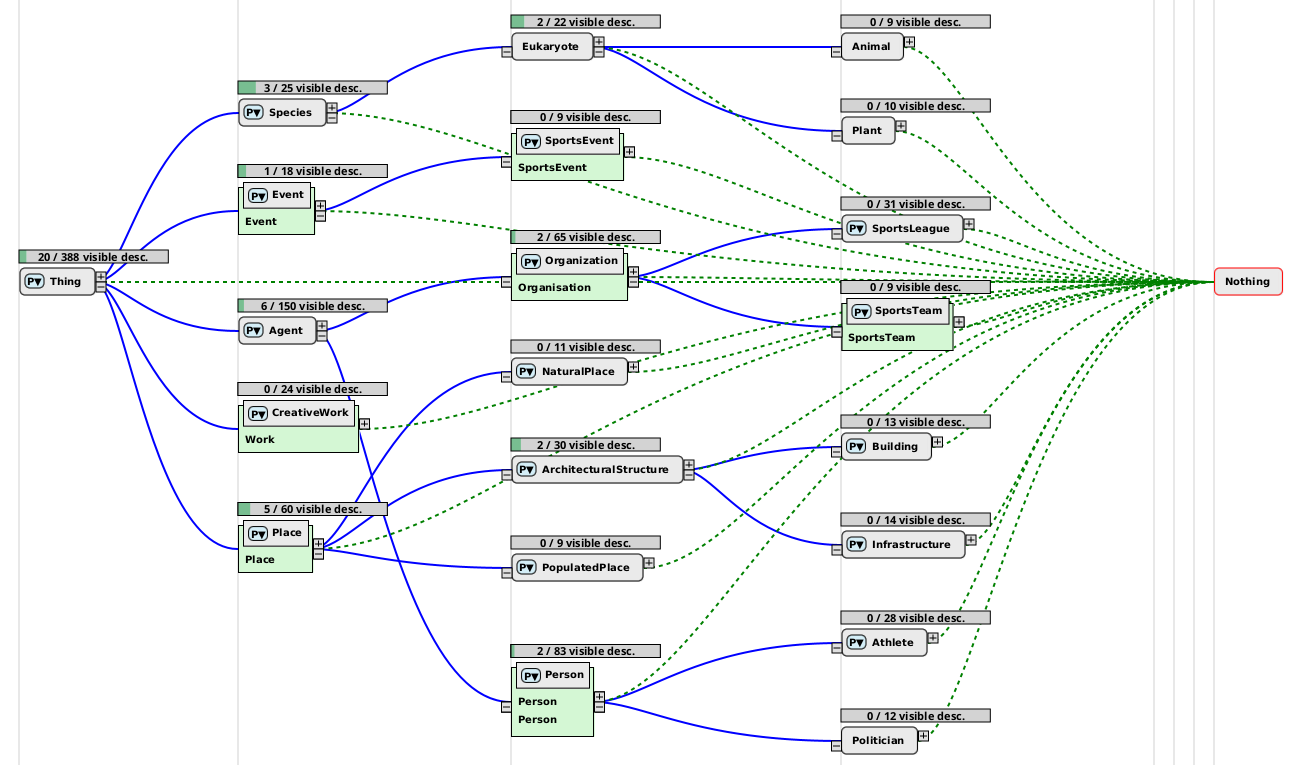}
        \caption{}
        \label{fig:ontviewDBP}
    \end{subfigure}
    ~
    \begin{subfigure}[c]{0.25\textwidth}
        \centering
        \includegraphics[width=\textwidth]{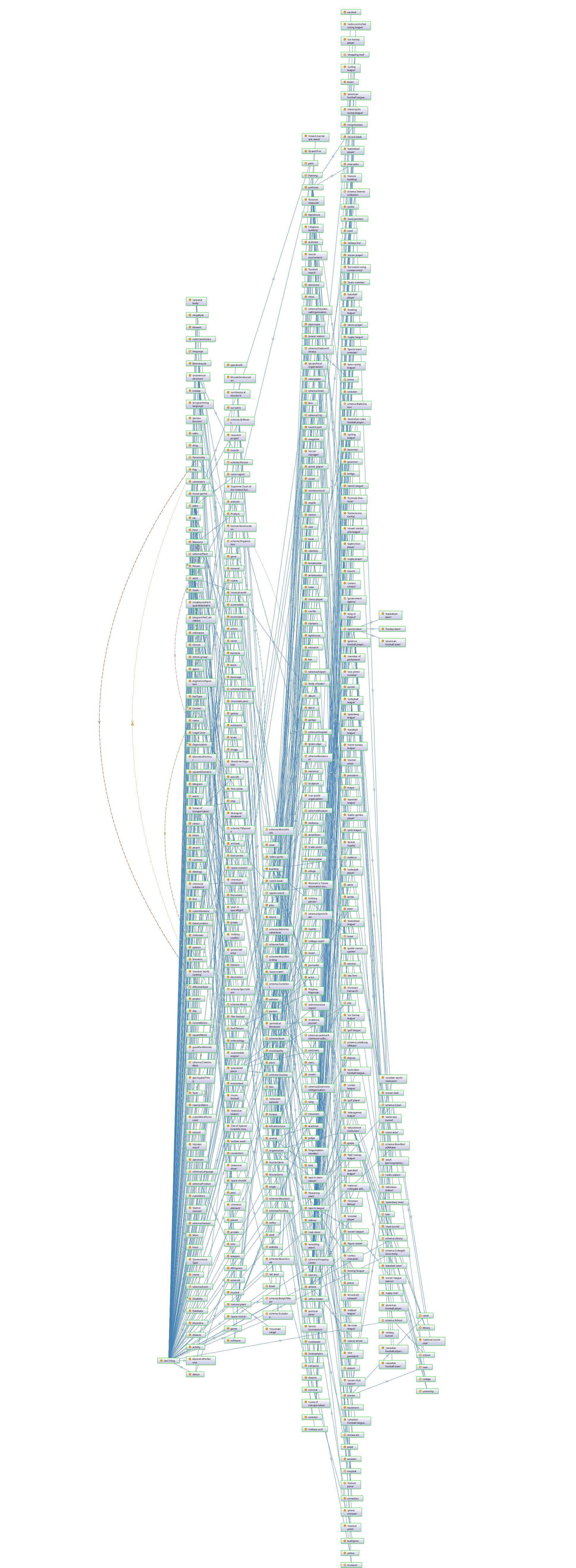}
        \caption{}
        \label{fig:ontogrfDBP}
    \end{subfigure}
    \caption{DBpedia ontology in a) OntView (20 node summary), b) Ontograf}
    \label{fig:dbp}
\end{figure}

Figure~\ref{fig:dbp} provides a comparison between OntView and Ontograf, illustrating how the DBpedia\_3.8 core ontology is rendered. In OntView, a PageRank–based summarization technique is applied to limit the display to the $20$ most relevant nodes, making it easier to identify key relationships among concepts. In contrast, Ontograf does not offer this capability, rendering the entire ontology and making it difficult for the user to clearly discern the most important node relationships.

\paragraph{Controlling the Expansion.} 

When dealing with very large ontologies, for instance, when concept hierarchies are unbalanced in terms of the number of descendants or the number of concepts with definitions, 
users need some kind of control over the expansion and collapse of the nodes. Otherwise, 
showing or hiding all the descendants at once would defeat the purpose of helping users to understand the ontology. 

\begin{figure}[!htb]
    \centering
       \begin{subfigure}[c]{0.22\textwidth}
        \centering
        \includegraphics[width=\textwidth]{Assets/toolbar/visibilityStep.png}
        \caption{}
        \label{fig:visibilityStepAgain}
    \end{subfigure}%
    ~
    \begin{subfigure}[c]{0.5\textwidth}
        \centering
        \includegraphics[width=\textwidth]{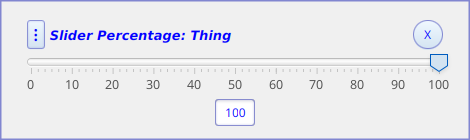}
        \caption{}
        \label{fig:sliderPercentage}
    \end{subfigure}
    \caption{Expansion Control: a) Percentage applied to each expansion step, b) Slider to control the current percentage of visible descendants of a selected node.}
\end{figure}

To avoid this, OntView provides a method to control the steps of the expansion by stating the percentage of nodes that the user wants to show/hide in each step (see  Fig.~\ref{fig:visibilityStepAgain}). Note that this percentage is applied to the number of descendants of a local node (it is a global parameter applied locally to each node). The policy that OntView follows to select the nodes has also been implemented in a modular way. Currently, OntView provides three main possible policies: 

\begin{itemize}
\item Select the next step elements using a relevance measure to be shown/hidden by ordering them according to a particular available relevance measure\footnote{OntView currently uses a RDFRank, but any method that returns the relevance value of an ontological element could be used.} regardless of the hierarchy level of the descendants. 

\item Select the next step elements prioritizing the most general descendants; thus, filling/emptying the descendant levels from left to right. 

\item Select the next step elements prioritizing the most specific descendants; thus, filling/emptying the descendant levels from right to left. 
\end{itemize}

The latter two policies are also guided by the relevance measure (i.e., the nodes within the same hierarchical level, are selected ordered by their relevance) and are applied directly only to the descendants of a particular node. In the case of ancestors, there is another precondition: they can only be hidden if they do not have any other visible descendants. 

\begin{figure}[!htb]
    \centering

    
    \begin{subfigure}[c]{\textwidth}
        \centering
        \includegraphics[width=\textwidth]{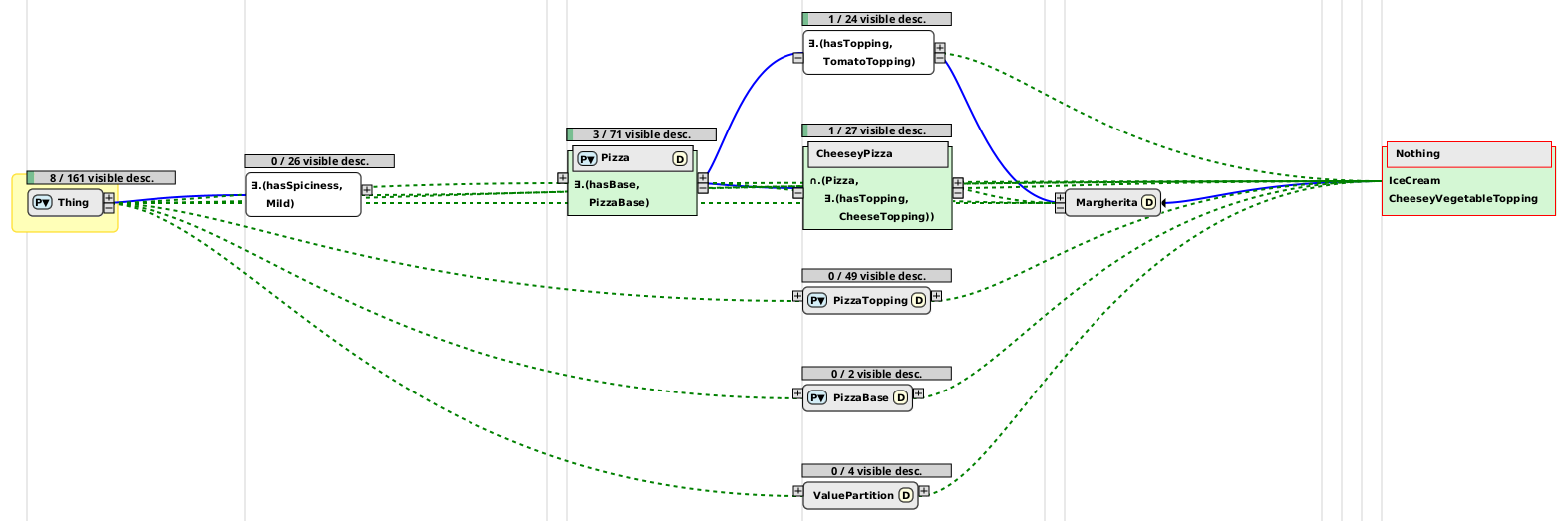}
        \caption{}
        \label{fig:stateA}
    \end{subfigure}
    
    \begin{subfigure}[c]{\textwidth}
        \centering
        \includegraphics[width=\textwidth]{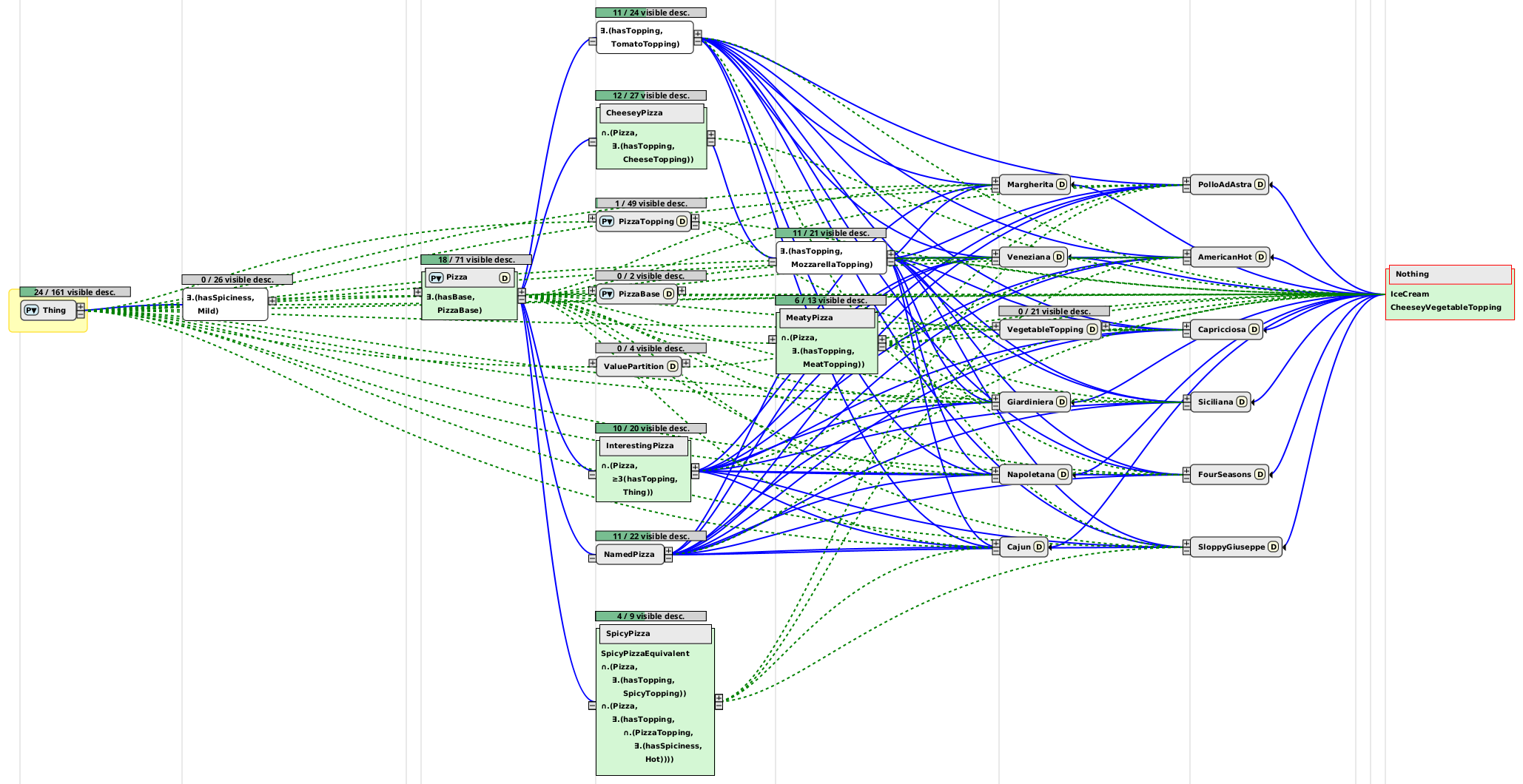}
        \caption{}
        \label{fig:stateB}
    \end{subfigure}
    \caption{Eye-hawk overview of a) 5\%, and b) 15\% nodes view having selected \texttt{owl:Thing} with Pizza ontology. Using this method allows the user to see \emph{emerging} ontology areas where the ontologies are finer-grained developed (e.g., helping to detect subdomains).}
    \label{fig:states}
\end{figure}

Moreover, to further ease exploration understanding, apart from expanding in a step-wise way, OntView also allows users to control a node expansion with a sliding bar\footnote{Following the same node selection policies as for the step-wise expansion.} (by right‐clicking on a node and selecting ``Show Percentage Visibility'' option), as illustrated in  Fig.~\ref{fig:sliderPercentage}. In fact, it provides the user with an extra summarization method: When selecting \texttt{owl:Thing}, OntView allows to expand the ontology by globally showing/hiding the most relevant concepts using this bar as it can be seen in Figures~\ref{fig:stateA}, and~\ref{fig:stateB}. In this case, we advocate for selecting the global relevance measure policy to get a broad view of the ontology (i.e., expanding the most general nodes in the hierarchy, such as \texttt{owl:Thing}) and the level-based strategies to gain further grain control as we delve into more specific branches of the ontology. 
%

\section{Prototype: Implementation Notes} 
\label{sec:evaluation}

OntView has been developed using Java as programming language using: 1)~the OWLAPI~\cite{horridge2011owl} as an API to handle the ontologies and as an interface for the different Description Logic reasoners we have used, and 2)~JavaFX as a graphical API. OntView has endured several changes of Java and OWLAPI versions, currently we are working with Java~17 and OWLAPI~5.1.9. However, we must state that we have witnessed a decay in the amount and availability of DL reasoners, which, back in the day, were a lively ecosystem. Using the OWLAPI~5.*, we decided to use Openllet (a fork of Pellet~2.0~\cite{pellet}, which is maintained by an active community)\footnote{\url{https://github.com/Galigator/openllet}, last accessed May 14, 2025.} and HermiT~\cite{hermit}.\footnote{We want to thank their contributors and maintainers for keeping them updated and alive.}


Much of the graphical development has been implemented by our research team, but for the initial placement of nodes, we used the implementation provided by pedviz API\footnote{\url{https://github.com/lukfor/pedviz/}, last accessed May 14, 2025.} of Sugiyama's algorithm to minimize the number of connectors crosses.
The source code of the current version of OntView can be downloaded from our github repository\footnote{\url{https://github.com/cbobed/OntView}, last accessed June 20, 2025.}. 



\section{Conclusions and Future Work}
\label{sec:conclusions}

In this paper, we have presented the ontology viewer OntView, an intuitive yet powerful tool for both ontology creators and users, and has been released with an open-source license.

OntView offers features such as observing the class and properties hierarchies using a level-based view. Despite using an expressive visual language to present the different ontology elements, it is able to display not only named classes but also anonymous ones (appearing on complex class expressions) and General Concept Inclusion (GCI) axioms, being to the best of our knowledge the first ontology viewer to do so. Furthermore, the use of a Description Logic reasoner ensures that inferred knowledge is also visualized.

Another notable feature is the ability to summarize the ontological knowledge according to different algorithms, and to modify the layout of the different visual elements dynamically as a result of user interactions. In this regard, our preliminary experiments, reported as supplementary material, show that OntView is the most appropriate tool to visualize and navigate ontologies which are large in size and complex. For example, DBpedia ontology can be displayed in less than $10$ seconds.

As future work, we would like to perform a systematic evaluation of our tool when used by different kinds of users, to clarify the pros and cons of existing ontology viewers in different scenarios. Furthermore, we would like to extend OntView with further functionalities. For example, given an individual, we would like to make it possible to display its  classes and property assertions, considering that the ABox could be large in some cases (e.g., Linked Open Data ontologies). Finally, we are considering to develop a Protegé plugin version to broaden its adoption by the community.


\begin{credits}

\subsubsection{\discintname}

The authors have no competing interests to declare that are relevant to the content of this article. 


\subsubsection{Resource Availability Statement}

Source code for OntView is available from our github repository\footnote{\url{https://github.com/cbobed/OntView}, last accessed June 20, 2025.}. Documentation about OntView is available from its webpage\footnote{\url{https://sid.cps.unizar.es/projects/OntView/}, last accessed June 20, 2025.}.

\subsubsection{\ackname} 

We were partially supported by the I+D+i projects PID2020-113903RB-I00 (funded by MCIN/AEI/10.13039/501100011033) and T42\_23R (funded by Gobierno de Arag\'on). We would like to thank Álvaro Juan for his help with the implementation of a preliminar prototype.

\end{credits}

%
%
\bibliographystyle{splncs04}
\bibliography{bibliography}

\end{document}